\documentclass[preprint,12pt]{elsarticle}
\usepackage{amssymb}
\usepackage{amsmath}
\allowdisplaybreaks[4]

\usepackage{caption}
\usepackage{subcaption}
\usepackage{adjustbox}
\usepackage{array}
\usepackage{bm}  
\usepackage{url} 
\usepackage{seqsplit}

\DeclareMathOperator\diff{d\!}
\def\wildcard{\mathord{\,\cdot\,}}

\journal{Applied Soft Computing}

\begin{document}

\begin{frontmatter}

%% Title, authors and addresses

%% use the tnoteref command within \title for footnotes;
%% use the tnotetext command for theassociated footnote;
%% use the fnref command within \author or \affiliation for footnotes;
%% use the fntext command for theassociated footnote;
%% use the corref command within \author for corresponding author footnotes;
%% use the cortext command for theassociated footnote;
%% use the ead command for the email address,
%% and the form \ead[url] for the home page:
%% \title{Title\tnoteref{label1}}
%% \tnotetext[label1]{}
%% \author{Name\corref{cor1}\fnref{label2}}
%% \ead{email address}
%% \ead[url]{home page}
%% \fntext[label2]{}
%% \cortext[cor1]{}
%% \affiliation{organization={},
%%             addressline={},
%%             city={},
%%             postcode={},
%%             state={},
%%             country={}}
%% \fntext[label3]{}

\title{Time Series Clustering with General State Space Models \\ via Stochastic 
Variational Inference}

%% use optional labels to link authors explicitly to addresses:
%% \author[label1,label2]{}
%% \affiliation[label1]{organization={},
%%             addressline={},
%%             city={},
%%             postcode={},
%%             state={},
%%             country={}}
%%
%% \affiliation[label2]{organization={},
%%             addressline={},
%%             city={},
%%             postcode={},
%%             state={},
%%             country={}}

\author[1]{Ryoichi Ishizuka}
\ead{ryoichi-ishizuka@biwako.shiga-u.ac.jp}
\author[2]{Takashi Imai}
\ead{takashi-imai@biwako.shiga-u.ac.jp}
\author[2]{Kaoru Kawamoto}
\ead{kaoru-kawamoto@biwako.shiga-u.ac.jp}

%% Author affiliation
\affiliation[1]{organization={Data Science and AI Innovation Research Promotion 
Center},
 addressline={Shiga University}, 
 country={Japan}}

\affiliation[2]{organization={Department of Data Science},
 addressline={Shiga University}, 
 country={Japan}}

%% Abstract
\begin{abstract}
%% Text of abstract
In this paper, we propose a novel method of model-based time series clustering with mixtures of general state space models (MSSMs). Each component of MSSMs is associated with each cluster. An advantage of the proposed method is that it enables the use of time series models appropriate to the specific time series. This not only improves clustering and prediction accuracy but also enhances the interpretability of the estimated parameters. The parameters of the MSSMs are estimated using stochastic variational inference, a subtype of variational inference. The proposed method estimates the latent variables of an arbitrary state space model by using neural networks with a normalizing flow as a variational estimator. The number of clusters can be estimated using the Bayesian information criterion. In addition, to prevent MSSMs from converging to the local optimum, we propose several optimization tricks, including an additional penalty term called entropy annealing.
To our best knowledge, the proposed method is the first computationally feasible one for time series clustering based on general (possibly nonlinear, non-Gaussian) state space models.
Experiments on simulated datasets show that the proposed method is effective for clustering, parameter estimation, and estimating the number of clusters.
\end{abstract}

%%Graphical abstract
%\begin{graphicalabstract}
%\includegraphics{grabs}
%\end{graphicalabstract}

%%Research highlights
%\begin{highlights}
%\item We propose a novel time series clustering method with mixtures of general SSMs.

%\item We can easily interpret estimated parameters of tailored time series models.

%\item Parameter estimation of SSM mixtures with SVI is computationally feasible.

%\item Normalizing flows increase the expressiveness of variational state estimators.

%\item Several tricks to avoid locally optimal parameter values are also proposed.

%\end{highlights}

%% Keywords
\begin{keyword}
%% keywords here, in the form: keyword \sep keyword
Time series clustering \sep Model-based clustering \sep State space model \sep Stochastic variational inference \sep Neural network
%% PACS codes here, in the form: \PACS code \sep code 

%% MSC codes here, in the form: \MSC code \sep code
%% or \MSC[2008] code \sep code (2000 is the default)

\end{keyword}

\end{frontmatter}

%% Add \usepackage{lineno} before \begin{document} and uncomment 
%% following line to enable line numbers
%% \linenumbers

%% main text
%%

%% Use \section commands to start a section
\section{Introduction}

Time series data analyses have been conducted in various fields, including science, engineering, business, finance, economics, medicine, and politics \citep{rani2012recent, aghabozorgi2015timeseries}. Time series clustering is an analysis method used to classify multiple time series data into groups with identical patterns. This technique is crucial for work in various fields, including recognizing pathological patterns using electrocardiogram data in medicine \cite{he2017automatic}, analyzing industry trends on acquisitions and restructuring in economics \cite{mitchell1996impact}, and monitoring condition of industrial machinery in engineering \cite{owsley1997automatic}.

There are three principal approaches to time series clustering: the shape-based approach, the feature-based approach, and the model-based approach \citep{aghabozorgi2015timeseries, warrenliao2005clustering}. The model-based approach has two advantages over the other approaches. First, it often shows higher accuracy when the model can adequately represent the pattern of the time series being analyzed. Second, it allows us to make predictions using the estimated model \citep{umatani2023time}. Thus, the model-based approach is particularly effective if an appropriate predictive model is used.

To ensure accurate clustering and predictions, it is important that the time series model adequately describes the dynamics of the time series. Typical examples in previous studies are an autoregressive (AR) model \citep{xiong2004time, venkataramanakini2013bayesian}, a hidden Markov model (HMM) \citep{li2000bayesian}, and a linear Gaussian state space model (LGSSM) \citep{umatani2023time}. However, the AR model has the limitation that it cannot adequately describe non-stationary time series. In addition, it is difficult to understand the underlying dynamics through the estimated AR model. Although the HMM has latent variables that allow us to construct a rich class of models, its application is limited to cases in which the latent variables are discrete. The LGSSM, in contrast, is capable of handling continuous latent variables and non-stationary time series. However, it is limited in its ability to accurately represent nonlinear and non-Gaussian dynamics.

In the present work, we propose a novel method for model-based time series clustering with general state space models \citep{kitagawa1987nongaussian, tanizaki1998nonlinear}, which allows the use of arbitrarily state and observation equations. An advantage of the proposed method is the availability of highly expressive time series models specific to the time series. This means that we can explicitly incorporate prior knowledge of the time series into the time series model. This improves clustering and forecasting accuracy while additionally contributing to the interpretability of the estimated parameters. The proposed method relies on the idea of finite mixture models \citep{mclachlan1988mixture} to introduce mixtures of state space models (MSSMs). The method classifies time series datasets into finite groups (clusters), and simultaneously estimates the model parameters corresponding to each cluster. The number of clusters can be estimated using the Bayesian information criterion (BIC).

The MSSMs are trained using stochastic variational inference (SVI) \citep{hoffman2013stochastic}, a subtype of variational inference (VI) \citep{bishop2006pattern, beal2003variational}. Variational inference is a method of approximating a complex posterior distribution with a tractable distribution, where the approximated distribution should be highly expressive. With SVI, the expressive power of the approximate distribution can be improved by using a neural network in its construction. The proposed method uses normalizing flows \citep{rezende2015variational, papamakarios2021normalizing} as the core of the variational estimator neural network, which further increases the expressive power of the approximate distribution.

SVI is a scalable solution to parameter estimation in state space models because it can exploit parallel computation in the time dimension. The proposed SVI-based method is, to our best knowledge, the first computationally feasible one for time series clustering based on general state space models.

This paper also proposes several optimization tricks to prevent the convergence of the approximate distribution to a local optimum in the estimation of MSSMs, the main one of which is introducing the penalizing technique called entropy annealing to the training of the cluster estimator. This trick contributes to the stability of parameter estimation.

The remainder of the paper is organized as follows. We summarize related work in Sec.~2, and review the SVI approach to parameter estimation, particularly using normalizing flows, in Sec.~3. The SVI approach is extended to MSSMs in Sec.~4. We demonstrate the effectiveness of this method via experiments on simulated datasets in Sec.~5.

\section{Related Work}

Recently, many model-based time series clustering methods have been presented on the idea of finite mixtures \citep{mclachlan1988mixture} of time series models. Some methods, particularly those based on mixtures of AR models \cite{xiong2004time, venkataramanakini2013bayesian}, HMMs \cite{li2000bayesian}, and LGSSMs \cite{umatani2023time}, have already achieved promising performance in terms of computational cost. However, no efficient methods have been proposed for mixtures of general state space models (that is, MSSMs). Although a possible candidate is the Markov chain Monte Carlo (MCMC)--based adaptable method proposed in \cite{frohwirth-schnatter2008modelbased}, its extension to MSSMs would result in a computationally intensive method.

This paper proposes a parameter estimation method based on SVI, instead of MCMC, for MSSMs. SVI can estimate parameters for complex posterior distributions for which analytical solutions cannot be computed, and is computationally less expensive than MCMC \citep{kingma2016improved}. Recently, SVI has been widely substituted for MCMC methods due to SVI's superior computational efficiency. See, for example, \cite{NIPS2014_865dfbde, povala2022variational, xuan2023stochastic}.

\section{Parameter Estimation using SVI}

\subsection{Variational Inference}

Variational inference (VI) approximates the posterior distribution of latent variables by a tractable probability distribution parameterized by $\phi$, and optimizes $\phi$ such that this distribution is closest to the posterior \citep{bishop2006pattern, beal2003variational}. Specifically, $\phi$ is estimated by minimizing the KL divergence between the approximate and posterior distributions. Let $\mathbf{y}$ denote the observed data, $\mathbf{x}$ the latent variables, and $\theta$ the parameter of the posterior distribution. The KL divergence of the posterior $p_{\theta} (\mathbf{x}\mid \mathbf{y})$ and the approximate distribution $q_{\phi} (\mathbf {x}\mid \mathbf{y})$ can be expressed as
\begin{align}
&D_{\mathrm{KL}}
            \left(q_{\phi}(\mathbf{x} \mid \mathbf{y}) 
                \parallel  p_{\theta}(\mathbf{x} \mid \mathbf{y})\right)
    =\int q_{\phi}(\mathbf{x} \mid \mathbf{y})
        \log  
            \frac{q_{\phi}(\mathbf{x} \mid \mathbf{y})}
                 {p_{\theta}(\mathbf{x} \mid \mathbf{y})}
                    \diff \mathbf{x} \notag\\ 
    &=\int q_{\phi}(\mathbf{x} \mid \mathbf{y})
        \log  
            \frac{q_{\phi}(\mathbf{x} \mid \mathbf{y})p_{\theta}(\mathbf{y})}
                 {p_{\theta}(\mathbf{x},  \mathbf{y})}
                    \diff \mathbf{x} \notag\\ 
    &=\int q_{\phi}(\mathbf{x} \mid \mathbf{y})
        \log p_{\theta}(\mathbf{y})\diff \mathbf{x}
        +\int q_{\phi}(\mathbf{x} \mid \mathbf{y})
        \log
            \frac{q_{\phi}(\mathbf{x} \mid \mathbf{y})}
                 {p_{\theta}(\mathbf{x},  \mathbf{y})}
                    \diff \mathbf{x} \notag\\ 
    &=\log p_{\theta}(\mathbf{y})-\text{ELBO}, 
\end{align}
where
\begin{align}
        \text{ELBO}
            &=\int q_{\phi}(\mathbf{x} \mid \mathbf{y})
            \log
                \frac{p_{\theta}(\mathbf{x},  \mathbf{y})}
                     {q_{\phi}(\mathbf{x} \mid \mathbf{y})}
                        \diff \mathbf{x} \notag\\
            &=E_{q_{\phi}(\mathbf{x} \mid \mathbf{y})} 
                \left[f_{(\phi, \theta)}(\mathbf{x})
                \right],  \\[.5ex]
            f_{(\phi, \theta)}(\mathbf{x})
            &=-\log q_{\phi}(\mathbf{x} \mid \mathbf{y})+\log p_{\theta}(\mathbf{x},  \mathbf{y}). 
\end{align}

From the above, the KL divergence is the difference between the marginal log-likelihood $\log p_\theta(\mathbf{y})$ and the lower bound of the marginal log-likelihood, or evidence lower bound (ELBO). When the KL divergence vanishes, the marginal log-likelihood and ELBO become identical. Thus, VI maximizes the ELBO with respect to $\phi$ and $\theta$ to achieve minimization of the KL divergence and maximization of the marginal log-likelihood.

\subsection{Stochastic Variational Inference and Re-parameterization Trick}

To obtain a good approximation of the posterior distribution of latent variables in VI, the approximate distribution should be as expressive as possible. Therefore, a method has been proposed to increase the expressive power of the approximate distribution by exploiting the high expressive power of neural networks \citep{kingma2013autoencoding, rezende2015variational}. The ELBO is maximized by stochastic gradient decent with respect to $\theta$ and $\phi$, and this method is called stochastic variational inference (SVI) \citep{hoffman2013stochastic}.

A re-parameterization trick was proposed in \cite{kingma2013autoencoding} as a method for computing the ELBO gradient with low variance. This trick expresses $\mathbf{x}$ as the deterministic function $g_{\phi}(\mathbf{\epsilon}, \mathbf{y})$ with the random vector $\epsilon$. Using this trick, the ELBO gradient is calculated as
\begin{align}
    \nabla_{\phi,\theta} \text{ELBO} \notag
        &=\nabla_{\phi,\theta} E_{p(\mathbf{\epsilon})} 
            \left[
            f_{\phi, \theta}
            \left(
            g_{\phi}(\mathbf{\epsilon}, \mathbf{y})
            \right)
            \right]  \notag\\ 
        &= E_{p(\mathbf{\epsilon})} 
            \left[\nabla_{\phi,\theta}
            \left(
            f_{\phi, \theta}
            \left(g_{\phi}(\mathbf{\epsilon}, \mathbf{y})
            \right)
            \right)   
            \right] \notag \\ 
        &\approx
            \frac{1}{L}\sum_{l=1}^{L}
            \left[\nabla_{\phi,\theta}
            \left(
            f_{\phi, \theta}
            \left(
                g_{\phi}(\mathbf{\epsilon}^{(l)}, \mathbf{y})
            \right)
            \right)  
            \right], 
\end{align}
where $\mathbf{\epsilon}^{(l)}\sim p(\mathbf{\epsilon})$, and $L$ is the number of Monte Carlo samples. This trick allows $\epsilon$ to be sampled independently of $\phi$, and thus $\theta$ and $\phi$ to be optimized by gradient decent methods.

\subsection{Normalizing Flows}

Normalizing flows \citep{rezende2015variational, papamakarios2021normalizing} are a tool for transforming random variables using continuous invertible functions, which is useful for improving the expressive power of the approximate distribution $q_\phi(\mathbf{x} \mid \mathbf{y})$. The likelihood of the random variables transformed by $f:\mathbf{\xi} \to\mathbf{\xi'}$ can be calculated from the Jacobian property of the invertible function as
\begin{align}
    p(\mathbf{\xi'}) 
        = p(\mathbf{\xi}) \left| \det \left( \frac{\partial f}{\partial \mathbf{\xi}} \right)^{-1}\right|. 
\end{align}

Successive application of normalizing flows as $f = f_F \circ \cdots \circ f_1$ produces more complex distributions. The log-likelihood through the normalizing flows is calculated as
\begin{align}
    \log p(\mathbf{\xi}_{F}) 
        &= \log p( \mathbf{\xi}) 
            + \sum_{f=1}^{F} \log \left| \det \left( \frac{\partial f_f}{\partial \mathbf{\xi}_{f-1}} \right)^{-1} \right|,
\end{align}
where $\mathbf{x} = \mathbf{\xi}_{F}$ and $\mathbf{\xi} = \mathbf{\xi}_0$.

In SVI, the function $f$ can be a neural network, which incorporates the normalizing flows as part of the approximate distribution parameterized by $\phi$, providing the approximate distribution with high expressive power.

\section{Proposed Method}
\subsection{Mixtures of State Space Models}

The state space model (SSM) \citep{kitagawa1987nongaussian, tanizaki1998nonlinear} is defined as
\begin{subequations}
\begin{align}
    \mathbf{x}{[t]} &\sim Q (\wildcard \mid \mathbf{x}{[t-1]}),\\
    \mathbf{y}{[t]} &\sim R (\wildcard \mid \mathbf{x}{[t]}),\\
    \mathbf{x}{[1]} &\sim P_0(\wildcard),
\end{align}
\end{subequations}
where $\mathbf{Y} = \{\mathbf{y}{[t]}\}_{t=1}^T$ and $\mathbf{X} = \{\mathbf{x}{[t]}\}_{t=1}^T$ are observed and latent variables, $Q$ denotes the conditional density function of $\mathbf{x}{[t]}$ given $\mathbf{x}{[t-1]}$, $R$ denotes the conditional density function of $\mathbf{y}{[t]}$ given $\mathbf{x}{[t]}$, and $P_0$ denotes the density function of the initial state $\mathbf{x}{[1]}$. The SSM then has parameters $\mathbf{\theta}=\{\mathbf{\theta}_Q, \mathbf{\theta}_R, \mathbf{\theta}_{P_{0}}\}$, where $\mathbf{\theta}_Q$, $\mathbf{\theta}_R$, and $\mathbf{\theta}_{P_{0}}$ are the parameters of $Q$, $R$, and $P_0$, respectively.

We can define mixtures of SSMs (MSSMs) of the above form. Let $D_\mathbf{Y}=\{\mathbf{Y}_i\}_{i=1}^{N}$ be a dataset consisting of $N$ observation time series $\mathbf{Y}_i = \{\mathbf{y}_{i}[t]\}_{t=1}^T$, and for $k \in \{1, 2, 3, \ldots, M\}$
\begin{subequations}
\begin{align}
    \mathbf{x}_{i}[t] &\sim Q^{(k)} (\wildcard \mid \mathbf{x}_{i}[t-1]),\\
    \mathbf{y}_{i}[t] &\sim R^{(k)} (\wildcard \mid \mathbf{x}_{i}[t]),\\
    \mathbf{x}_{i}[1] &\sim P_0^{(k)}(\wildcard)
\end{align}
\end{subequations}
be different SSMs, and define an MSSM as 
\begin{equation}
    p_\Theta({\mathbf{Y}_i,{\mathbf{X}_i}})
        =\sum_{k=1}^M p_{\theta^{(k)}}({\mathbf{Y}_i,{\mathbf{X}}_i})p^{(k)},
\end{equation}
where $\theta^{(k)}=\{\theta^{(k)}_Q,\theta^{(k)}_R,\theta^{(k)}_{P_0}\}$ is the parameters of the $k$-th SSM, $p^{(k)}$ is the weight of the $k$-th mixture component, and $\Theta=\{\theta^{(1)}, p^{(1)}, \theta^{(2)}, p^{(2)},\ldots, \theta^{(M)}, p^{(M)}\}$ is the whole parameter set of the MSSM. We can interpret each SSM as a cluster. Let $\mathbf{z} = \{z_i\}_{i=1}^N$ be the new latent variables, in which $z_i$ indicates the cluster of  $\mathbf{Y}_i$.

\subsection{SVI for MSSMs}

Introducing the new latent variables $\mathbf{z}$, the KL divergence $D_{\mathrm{KL}}\left(q_{\phi}(\mathbf{X}_i, z_i \mid \mathbf{Y}_i)\parallel p_{\Theta}(\mathbf{X}_i, z_i \mid \mathbf{Y}_i)\right)$
becomes
\begin{align}
&D_{\mathrm{KL}}
            \left(q_{\phi}(\mathbf{X}_i, z_i \mid \mathbf{Y}_i) 
                \parallel p_{\Theta}(\mathbf{X}_i, z_i \mid \mathbf{Y}_i)\right) \notag\\ 
    &=\log p_{\Theta}(\mathbf{Y}_i)-\text{ELBO}_i, 
    \label{eq:DKL_MSSM}
\end{align}
where
\begin{align}
        \text{ELBO}_i
            ={}&E_{q_{\phi}(\mathbf{X}_i, z_i \mid \mathbf{Y}_i)} 
                \left[
                    f_{ \Theta, \phi}(\mathbf{X}_i, z_i)
                \right],
           \\[.5ex]
        f_{ \Theta, \phi}(\mathbf{X}_i, z_i)
            ={}&-\log q_{\phi}(\mathbf{X}_i \mid z_i, \mathbf{Y}_i)
            -\log q_{\phi}(z_i \mid \mathbf{Y}_i) \notag\\
            &{}+\log p_{\Theta}(\mathbf{X}_i, z_i, \mathbf{Y}_i).
\end{align}
%The parameters are estimated by maximizing $\text{ELBO}_i$ for $\Theta$ and $\phi$. 
Assuming that $\mathbf{Y}_i$ are independent of each other, the log-likelihood, $\log p_{\Theta}(D_\mathbf{Y})$, of the dataset can be rewritten using Eq.~\eqref{eq:DKL_MSSM} as
\begin{align}
&\log p_{\Theta}(D_\mathbf{Y}) 
    = \sum_{i=1}^N \log p_{\Theta}(\mathbf{Y}_i) \notag\\
    &= \sum_{i=1}^N 
        \left[ \text{ELBO}_i
        + D_{\mathrm{KL}}
            \left(q_{\phi}(\mathbf{X}_i, z_i \mid \mathbf{Y}_i) 
                \parallel p_{\Theta}(\mathbf{X}_i, z_i \mid \mathbf{Y}_i)\right)
        \right].
\end{align}
Maximizing the sum of the ELBOs with respect to $\phi$ and $\Theta$ leads to minimization of the KL divergence and maximization of the marginal log-likelihood. In SVI, this sum is optimized by stochastic gradient decent with the loss function 
\begin{align}
    \mathcal{L}(\Theta, \phi \mid D_\mathbf{Y}) 
        = \frac{1}{N}\sum_{i=1}^{N}\mathcal{L}(\Theta, \phi \mid \mathbf{Y}_i) 
        = -\frac{1}{N}\sum_{i=1}^{N}\text{ELBO}_i.
\end{align}

In the proposed method, to increase the expressive power of the approximate distributions $q_{\phi}(\mathbf{X}_i \mid z_i, \mathbf{Y}_i)$ and $q_{\phi}(z_i \mid \mathbf{Y}_i)$, neural networks parameterized by $\phi$ are employed. Specifically, $q_{\phi}(\mathbf{X}_i \mid z_i, \mathbf{Y}_i)$ is a neural network containing normalizing flows, and $q_{\phi}(z_i \mid \mathbf{Y}_i)$ is a neural network with a softmax activation function in the final layer.

\begin{figure}[t]
    \centering
    \includegraphics[width=0.7\linewidth]{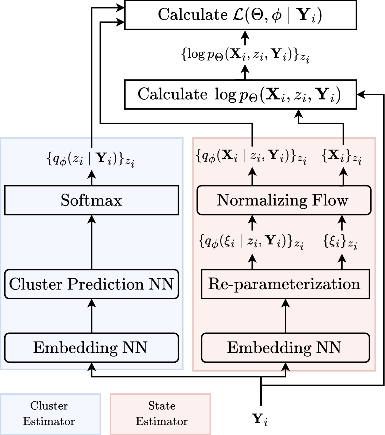}
    \caption{Overview of proposed method.}
    \label{fig:overview}
\end{figure}

By applying the re-parameterization trick, $\nabla_{\phi,\Theta} \text{ELBO}_i$ can be approximated as
\begin{align}
    &\nabla_{(\phi,\Theta)} \text{ELBO}_i \notag\\
        &=\nabla_{(\phi,\Theta)} 
        E_{p(\mathbf{\epsilon})q_{\phi}(z_i \mid \mathbf{Y}_i)} 
            \left[
            f_{(\phi,\Theta)}(
                g_{\phi}(\mathbf{\epsilon}, z_i, 
                \mathbf{Y}_i), z_i)
            \right] \notag \\ 
        &=E_{p(\mathbf{\epsilon})}
            \left[\nabla_{(\phi,\Theta)} 
            \sum_{z_i=1}^{M}q_{\phi}(z_i \mid \mathbf{Y}_i)
            f_{(\phi,\Theta)}(
                g_{\phi}(\mathbf{\epsilon}, z_i, 
                \mathbf{Y}_i), z_i)
            \right]  \notag\\ 
        &\approx
            \frac{1}{L}\sum_{l=1}^{L}
            \nabla_{(\phi,\Theta)} 
            \sum_{z_i=1}^{M}q_{\phi}(z_i \mid \mathbf{Y}_i)
            f_{(\phi,\Theta)}(
                g_{\phi}(\mathbf{\epsilon}, z_i, 
                \mathbf{Y}_i), z_i)  \notag\\
        &=\nabla_{(\phi,\Theta)} 
            \frac{1}{L}
            \left[
            \sum_{l=1}^{L}
            \sum_{z_i=1}^{M}q_{\phi}(z_i \mid \mathbf{Y}_i)
            f_{(\phi,\Theta)}(
                g_{\phi}(\mathbf{\epsilon}, z_i, 
                \mathbf{Y}_i), z_i)
            \right] ,
\end{align}
where $\mathbf{\epsilon}^{(l)}\sim p(\mathbf{\epsilon})\notag$. As in \cite{kingma2013autoencoding}, the number $L$ of Monte Carlo samples is 1 and $p(\mathbf{\epsilon})$ is the standard normal distribution in this paper. An overview of the proposed method is shown in Figure~\ref{fig:overview}.

Our model is similar to that proposed in \cite{pires2020variational}, but differs significantly in that it involves latent state variables. Another difference is that the weights $p^{(k)}$ in our model are also parameters to be estimated. These differences require additional tricks in parameter estimation, as described later.

\subsection{Estimating the Number of Clusters}
The number of clusters is estimated using the Bayesian information criterion (BIC) as in \cite{xiong2004time, li2000bayesian,umatani2023time}. The BIC is defined as 
\begin{align*}
   \text{BIC} = \log\, \mathrm{L} - \frac{1}{2} \bigl(\lVert P\rVert-1\bigr) \log\ N,
\end{align*}
and a model with a larger BIC value is preferred. Here, $\mathrm{L}$ is the likelihood of a model, $\lVert P \lVert$ is the number of parameters of the model, and $N$ is the number of data. In MSSMs, the mean of $\text{ELBO}_i$ is used as $\mathrm{L}$, and the number of parameters of $\Theta$ referred to as $\lVert P_{\Theta}\lVert$ is used as $ \lVert P \lVert$. Consequently, the BIC of the MSSMs is 
\begin{align}
    \label{eq:BIC}
    \text{BIC} = \frac{1}{N} \sum_{i=1}^N \text{ELBO}_{i}
        - \dfrac{1}{2} \bigl(\lVert P_{\Theta}\rVert-1\bigr) \log\, N .
\end{align}
If the number of clusters is unknown, we train models with different numbers of clusters and adopt the model that exhibits the largest BIC.

\subsection{Entropy Annealing}

In parameter estimation of MSSMs, $q_{\phi}(z_i \mid \mathbf{Y}_i)$ may concentrate on some clusters and converge to a local optimum in the early stages of training. To prevent this, we introduce an additional penalty term called entropy annealing. The loss function of the proposed method can be rewritten as
\begin{align}
 \mathcal{L}&(\Theta, \phi \mid D_\mathbf{Y}) = 
    -\frac{1}{N}\sum_{i=1}^N\text{ELBO}_i  \notag\\
    ={}
    &\frac{1}{N}\sum_{i=1}^N \left[-\mathcal{H}(q_{\phi} (z_i \mid \mathbf{Y}_i))
    -\sum_{z_i=1}^{M} q_{\phi} (z_i \mid \mathbf{Y}_i)\mathcal{H}(q_{\phi} (\mathbf{X}_i\mid z_i, \mathbf{Y}_i))  \right. \notag \\
    &\left.-\sum_{z_i=1}^{M}
    \int q_{\phi} (\mathbf{X}_i, z_i \mid \mathbf{Y}_i)
    \log p_{\theta^{(k)}} (\mathbf{X}_i,z_i, \mathbf{Y}_i)
    \diff \mathbf{X}_i \right], \label{eq:loss_analysis}
\end{align}
where $\mathcal{H}(q)$ indicates the entropy of $q$. From Equation (\ref{eq:loss_analysis}), the loss function comprises the entropy penalty term concerning the approximate distribution and the expected value of the complete data log-likelihood. Entropy annealing increases the entropy penalty $\mathcal{H}(q_{\phi} (z_i \mid \mathbf{Y}_i))$ in the early stages of training, which prevents $q_{\phi}(z_i \mid \mathbf{ Y}_i)$ from converging to the local optimum. The loss function with entropy annealing is
\begin{align} 
\mathcal{L}^{'}_{n}&(\Theta,\phi \mid \mathbf{Y}_i)
= \mathcal{L}(\Theta,\phi \mid \mathbf{Y}_i) -
{\alpha_{n}}\mathcal{H}(q_{\phi} (k | \mathbf{Y}_i)), 
\end{align}
where ${\alpha_{n}}$ is the strength of entropy annealing at epoch $n$. 

The strength ${\alpha_{n}}$ is dynamically changed by epoch as
\begin{align}
\alpha_n = \begin{cases} 
    \alpha & \text{if } 1 \leq n < n_{\alpha}^{\text{start}}, \\
    \frac{n_{\alpha}^{\text{end}} - n}{n_{\alpha}^{\text{end}} - n_{\alpha}^{\text{start}}} a_0 & \text{if } n_{\alpha}^{\text{start}} \leq n < n_{\alpha}^{\text{end}}, \\
    0 & \text{if } n_{\alpha}^{\text{end}} \leq n ,
\end{cases}
\end{align}
where $\alpha$, $n_{\alpha}^{\text{start}}$, and  $n_{\alpha}^{\text{end}}$ are the maximum annealing strength, number of epochs at which entropy annealing starts to weaken, and number of epochs at which entropy annealing ends, respectively. Other tricks to avoid convergence to a local optimum are described in Appendix A.

\section{Experiments}
{\footnotesize
\begin{table*}[tb]
    \captionsetup{justification=raggedright,singlelinecheck=false}
    \caption{Parameter values of Stuart--Landau oscillators.}
    \centering
    \begin{tabular}{cccccccccc} \hline
        Cluster & $a_1^{(k)}$ & $a_2^{(k)}$ & $b_1^{(k)}$ & $b_2^{(k)}$ & 
            $\mathbf{A}^{(k)}$ & $b^{(k)}$ & $\mu_{x}^{(k)}$ & $\mu_{y}^{(k)}$ &
            $\mathbf{C}^{(k)}$\\[1pt]\hline
        $1$ & $1.0$ & $0.5$ & $0.5$ & $0.1$ 
            & $0.05$ $\times \mathbf{I}_{2}$ 
            & $0.05$ & $1.0$ & $0.0$ 
            & $0.05$ $\times \mathbf{I}_{2}$ \\  % Content row 1
        $2$ & $1.0$ & $1.5$ & $0.8$ & $0.2$ 
            & $0.05$ $\times \mathbf{I}_{2}$ 
            & $0.05$ & $1.0$ & $0.0$
            & $0.05$ $\times \mathbf{I}_{2}$ \\  % Content row 2
        $3$ & $1.0$ & $1.0$ & $0.2$ & $0.0$ 
            & $0.05$ $\times \mathbf{I}_{2}$ 
            & $0.05$ & $1.0$ & $0.0$
            & $0.05$ $\times \mathbf{I}_{2}$ \\ \hline % Content row 3
    \end{tabular}
    \label{tab:sl_params}
\end{table*}
}
{\footnotesize
\begin{table*}[tb]
\centering
\captionsetup{justification=raggedright,singlelinecheck=false}
\caption{Results of experiment for Stuart--Landau oscillator dataset. (a) Five-trial mean BIC values for different values of $M$. The values in parentheses are standard deviations. (b) Total confusion matrix obtained from five trials. (c) Five-trial mean of estimated parameter values. The values in parentheses are standard deviations.}
\begin{subtable}{1\linewidth}
    \raggedright
    \centering
    \captionsetup{justification=raggedright,singlelinecheck=false}
    \caption{BIC values}
    \label{tab:sl_bic}
    \begin{tabular}{ccccc}\hline
    $M$ & 2 & 3 & 4 & 5\\ \hline
    BIC               & \begin{tabular}[c]{@{}c@{}}$-107.32$\\ ($7.60$)\end{tabular} 
                      & \begin{tabular}[c]{@{}c@{}}$25.83$\\ ($1.26$)\end{tabular} 
                      & \begin{tabular}[c]{@{}c@{}}$-17.55$\\ ($0.73$)\end{tabular} 
                      & \begin{tabular}[c]{@{}c@{}}$-61.85$\\ ($1.91$)\end{tabular} \\ \hline
    \end{tabular}
    \end{subtable}%
\vspace{0.5cm}
\begin{subtable}{1\linewidth}
    \raggedright
    \centering
    \captionsetup{justification=raggedright,singlelinecheck=false}
    \caption{Confusion matrix}
    \label{sl:confusion_matrix}
    \begin{tabular}{|l|c|c|c|} \hline
        True $\backslash$ Prediction & $1$ & $2$ & $3$ \\ \hline
        $1$ ($i = 001,\;\dotsc\,,\;300$) & 1500 & 0 & 0 \\ \hline
        $2$ ($i = 301,\;\dotsc\,,\;600$) & 0 & 1500 & 0 \\ \hline
        $3$ ($i = 601,\;\dotsc\,,\;900$) & 0 & 0 & 1500 \\ \hline 
    \end{tabular}
    \vspace{0.3cm}
\end{subtable}
\begin{subtable}{1.0\linewidth}
    \raggedright
    \captionsetup{justification=raggedright,singlelinecheck=false}
    \caption{Estimated parameter values}
    \label{tbl:sl_estmated_params}
    \footnotesize
    \scalebox{0.65}{
    \begin{tabular}{cccccccccccccc}
    \hline
    Cluster & $p^{(k)}$ & $a_2^{(k)}$ & $b_1^{(k)}$ & $b_2^{(k)}$ & $\mathbf{A}_{11}^{(k)}$ & $\mathbf{A}_{12}^{(k)}$ & $\mathbf{A}_{22}^{(k)}$ & $b^{(k)}$ & $\mathbf{C}_{11}^{(k)}$ & $\mathbf{C}_{12}^{(k)}$ & $\mathbf{C}_{22}^{(k)}$ & $\mu_x^{(k)}$ & $\mu_y^{(k)}$ \\[1pt] \hline
    1 & \begin{tabular}[c]{@{}c@{}}$0.334$\\ ($0.002$)\end{tabular} & \begin{tabular}[c]{@{}c@{}}$0.499$\\ ($0.005$)\end{tabular} & \begin{tabular}[c]{@{}c@{}}$0.497$\\ ($0.000$)\end{tabular} & \begin{tabular}[c]{@{}c@{}}$0.097$\\ ($0.000$)\end{tabular} & \begin{tabular}[c]{@{}c@{}}$0.077$\\ ($0.001$)\end{tabular} & \begin{tabular}[c]{@{}c@{}}$0.000$\\ ($0.001$)\end{tabular}  & \begin{tabular}[c]{@{}c@{}}$0.071$\\ ($0.001$)\end{tabular} & \begin{tabular}[c]{@{}c@{}}$0.026$\\ ($0.001$)\end{tabular} & \begin{tabular}[c]{@{}c@{}}$0.079$\\ ($0.008$)\end{tabular} & \begin{tabular}[c]{@{}c@{}}$0.018$\\ ($0.007$)\end{tabular}  & \begin{tabular}[c]{@{}c@{}}$0.060$\\ ($0.004$)\end{tabular} & \begin{tabular}[c]{@{}c@{}}$1.030$\\ ($0.016$)\end{tabular} & \begin{tabular}[c]{@{}c@{}}$0.006$\\ ($0.002$)\end{tabular}  \\ \hline
    2 & \begin{tabular}[c]{@{}c@{}}$0.333$\\ ($0.001$)\end{tabular} & \begin{tabular}[c]{@{}c@{}}$1.492$\\ ($0.002$)\end{tabular} & \begin{tabular}[c]{@{}c@{}}$0.795$\\ ($0.001$)\end{tabular} & \begin{tabular}[c]{@{}c@{}}$0.189$\\ ($0.001$)\end{tabular} & \begin{tabular}[c]{@{}c@{}}$0.084$\\ ($0.001$)\end{tabular} & \begin{tabular}[c]{@{}c@{}}$0.010$\\ ($0.001$)\end{tabular}  & \begin{tabular}[c]{@{}c@{}}$0.075$\\ ($0.001$)\end{tabular} & \begin{tabular}[c]{@{}c@{}}$0.021$\\ ($0.001$)\end{tabular} & \begin{tabular}[c]{@{}c@{}}$0.045$\\ ($0.006$)\end{tabular} & \begin{tabular}[c]{@{}c@{}}$-0.019$\\ ($0.003$)\end{tabular} & \begin{tabular}[c]{@{}c@{}}$0.073$\\ ($0.003$)\end{tabular} & \begin{tabular}[c]{@{}c@{}}$0.985$\\ ($0.007$)\end{tabular} & \begin{tabular}[c]{@{}c@{}}$-0.002$\\ ($0.002$)\end{tabular} \\ \hline
    3 & \begin{tabular}[c]{@{}c@{}}$0.333$\\ ($0.002$)\end{tabular} & \begin{tabular}[c]{@{}c@{}}$0.990$\\ ($0.000$)\end{tabular} & \begin{tabular}[c]{@{}c@{}}$0.201$\\ ($0.000$)\end{tabular} & \begin{tabular}[c]{@{}c@{}}$0.001$\\ ($0.000$)\end{tabular} & \begin{tabular}[c]{@{}c@{}}$0.082$\\ ($0.002$)\end{tabular} & \begin{tabular}[c]{@{}c@{}}$-0.002$\\ ($0.001$)\end{tabular} & \begin{tabular}[c]{@{}c@{}}$0.072$\\ ($0.001$)\end{tabular} & \begin{tabular}[c]{@{}c@{}}$0.023$\\ ($0.000$)\end{tabular} & \begin{tabular}[c]{@{}c@{}}$0.058$\\ ($0.008$)\end{tabular} & \begin{tabular}[c]{@{}c@{}}$-0.004$\\ ($0.008$)\end{tabular} & \begin{tabular}[c]{@{}c@{}}$0.069$\\ ($0.003$)\end{tabular} & \begin{tabular}[c]{@{}c@{}}$0.910$\\ ($0.086$)\end{tabular} & \begin{tabular}[c]{@{}c@{}}$0.002$\\ ($0.003$)\end{tabular}  \\ \hline
    \end{tabular}}
    \end{subtable}
    \label{tab:sl}
\end{table*}
}
\begin{figure*}[tb]
    \begin{minipage}[t]{0.5\linewidth}
        \includegraphics[width=\linewidth]{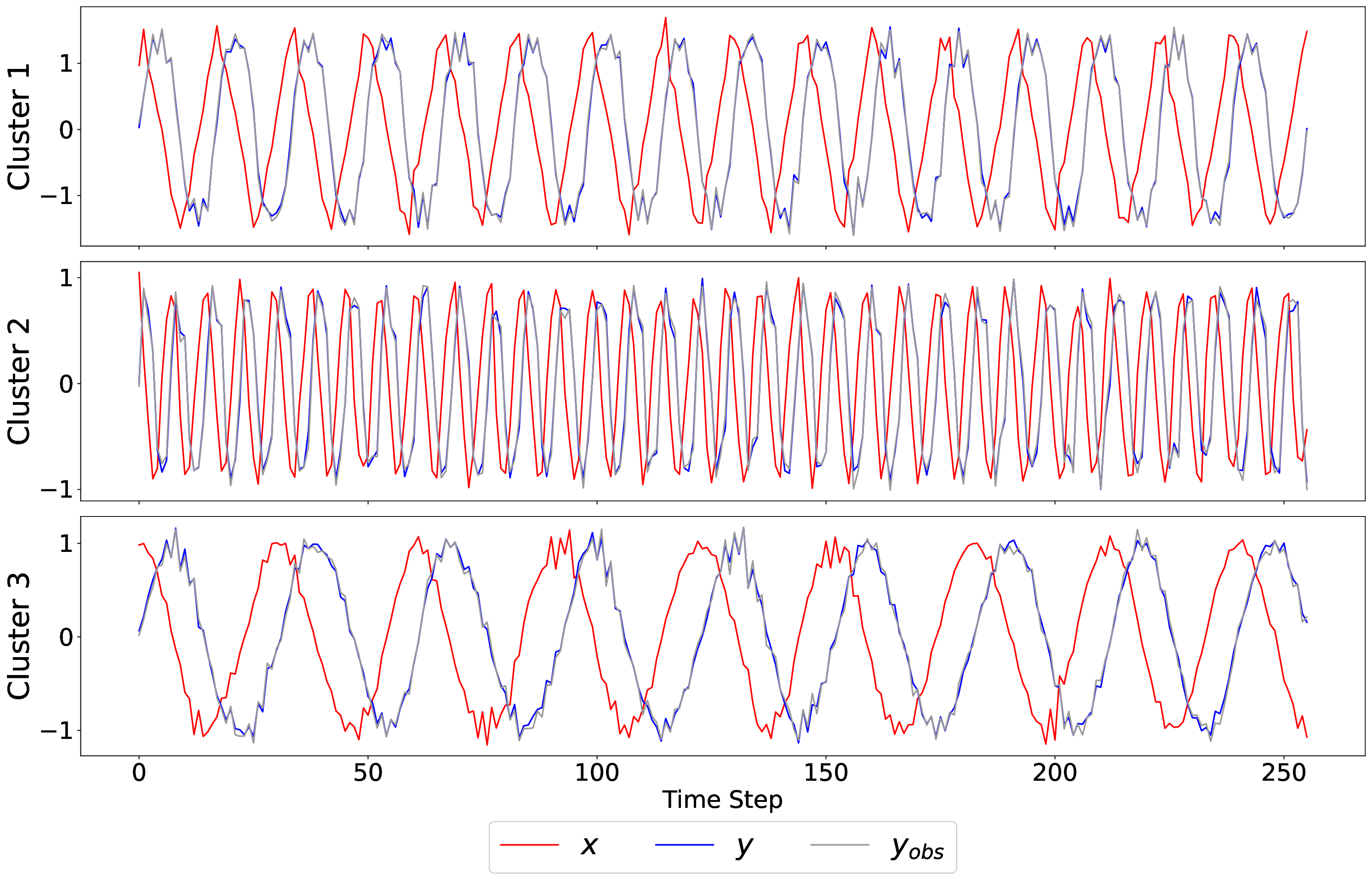}
        \subcaption{Actual time series}
        \label{fig:sl_actual}
    \end{minipage}
    \begin{minipage}[t]{0.5\linewidth}
        \includegraphics[width=\linewidth]{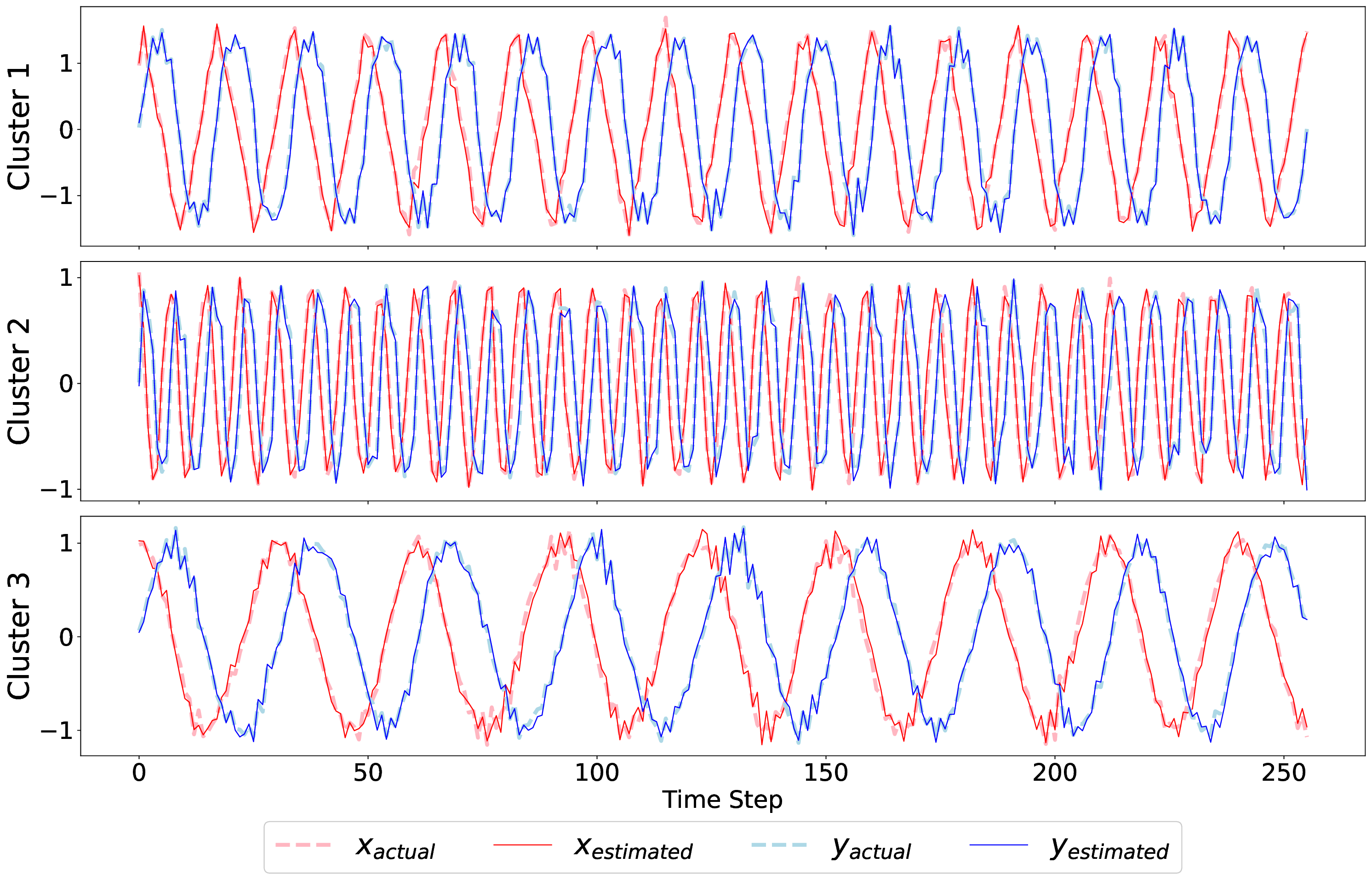}
        \subcaption{Estimated latent variables}
        \label{fig:sl_pred}
    \vspace{0cm}
    \end{minipage}
    \caption{Representative samples of the actual and estimated time series for the Stuart--Landau oscillator dataset. (a) Actual time series. The latent variable $y$ is observed as $y_\text{obs}$ under noise. (b) Estimated latent variables. For comparison, the actual latent variables are also plotted.}
    \label{fig:sl}
\end{figure*}

{\footnotesize
\begin{table}[tb]
    \captionsetup{justification=raggedright,singlelinecheck=false}
    \caption{Parameter values of SIR models.}
    \centering
    \begin{tabular}{ccccccccc} \hline
        Cluster & $\beta^{(k)}$ & $\gamma^{(k)}$ 
        & $a^{(k)}$ & $b^{(k)}$ & $c^{(k)}$
        & $\bm{\rho}_{\text({\text{init}})}^{(k)}$\\[2pt] \hline
        $1$ & $0.9$ & $0.1$ & $3$ & $3$ & $2$ 
        &$
            (0.95, 0.04, 0.01)
         $ \\ \hline % Content row 1
        2 & 0.3 & 0.05 & 3 & 3 & 2
        &$
            (0.95, 0.04, 0.01)
         $\\ \hline % Content row 2
    \end{tabular}
    \label{tab:sir_params} % Label for referencing
\end{table}
}
{\footnotesize
\begin{table*}[tb]
\centering
\captionsetup{justification=raggedright,singlelinecheck=false}
\caption{Results of experiment for SIR model dataset. (a) Five-trial mean BIC values for different values of $M$. The values in parentheses are standard deviations. (b) Total confusion matrix obtained from five trials. (c) Five-trial mean of estimated parameter values. The values in parentheses are standard deviations.}
\begin{subtable}[t]{1\linewidth}
    \raggedright
    \centering
    \captionsetup{justification=raggedright,singlelinecheck=false}
    \caption{BIC values}
    \label{tab:sir_bic}
    \begin{tabular}{ccccc}
    \hline
    Number of clusters & $2$ & $3$ & $4$ \\ \hline
    BIC               & \begin{tabular}[c]{@{}c@{}}$102.04$\\ ($0.36$)\end{tabular} 
                      & \begin{tabular}[c]{@{}c@{}}$71.30$\\ ($0.31$)\end{tabular} 
                      & \begin{tabular}[c]{@{}c@{}}$40.81$\\ ($0.30$)\end{tabular} \\ \hline
    \end{tabular}
    \end{subtable}%

\begin{subtable}[t]{1\linewidth}
    \raggedright
    \centering
    \captionsetup{justification=raggedright,singlelinecheck=false}
    \caption{Confusion matrix}
    \label{tab:sir_confusion_matrix}
    \begin{tabular}{|l|c|c|} \hline
        Cluster $\backslash$ Prediction & $1$ & $2$ \\ \hline
        $1$ ($i = 001,\;\dotsc\,,\;300$) & $1500$ & $0$ \\ \hline
        $2$ ($i = 301,\;\dotsc\,,\;600$) & $0$ & $1500$  \\ \hline
    \end{tabular}
    \vspace{0.3cm}
\end{subtable}
\begin{subtable}[b]{1.0\linewidth}
\centering
\captionsetup{justification=raggedright,singlelinecheck=false}
\caption{Estimated parameter values}
\scalebox{0.8}{
\begin{tabular}{c@{\extracolsep{\fill}}ccccccccc}
\hline
Cluster & $p^{(k)}$ & $\beta^{(k)}$ & $\gamma^{(k)}$ & $a^{(k)}$ & $b^{(k)}$ & $c^{(k)}$ & $\bm{\rho}_{(S, \text{init})}^{(k)}$ & $\bm{\rho}_{(I, \text{init})}^{(k)}$ & $\bm{\rho}_{(R, \text{init})}^{(k)}$ \\ [2pt]\hline
$1$      & \begin{tabular}[c]{@{}c@{}}$0.501$\\ ($0.002$)\end{tabular} & \begin{tabular}[c]{@{}c@{}}$0.851$\\ ($0.001$)\end{tabular} & \begin{tabular}[c]{@{}c@{}}$0.101$\\ ($0.000$)\end{tabular} & \begin{tabular}[c]{@{}c@{}}$2.801$\\ ($0.014$)\end{tabular} & \begin{tabular}[c]{@{}c@{}}$2.980$\\ ($0.010$)\end{tabular} & \begin{tabular}[c]{@{}c@{}}$2.686$\\ ($0.043$)\end{tabular} & \begin{tabular}[c]{@{}c@{}}$0.954$\\ ($0.002$)\end{tabular} & \begin{tabular}[c]{@{}c@{}}$0.037$\\ ($0.000$)\end{tabular} & \begin{tabular}[c]{@{}c@{}}$0.009$\\ ($0.002$)\end{tabular} \\ \hline
$2$      & \begin{tabular}[c]{@{}c@{}}$0.499$\\ ($0.002$)\end{tabular} & \begin{tabular}[c]{@{}c@{}}$0.293$\\ ($0.003$)\end{tabular} & \begin{tabular}[c]{@{}c@{}}$0.049$\\ ($0.001$)\end{tabular} & \begin{tabular}[c]{@{}c@{}}$2.763$\\ ($0.006$)\end{tabular} & \begin{tabular}[c]{@{}c@{}}$3.049$\\ ($0.007$)\end{tabular} & \begin{tabular}[c]{@{}c@{}}$2.715$\\ ($0.007$)\end{tabular} & \begin{tabular}[c]{@{}c@{}}$0.930$\\ ($0.013$)\end{tabular} & \begin{tabular}[c]{@{}c@{}}$0.037$\\ ($0.000$)\end{tabular} & \begin{tabular}[c]{@{}c@{}}$0.034$\\ ($0.013$)\end{tabular} \\ \hline
\end{tabular}}
\label{tab:sir_estmated_params}
\end{subtable}
\label{tab:sir}
\end{table*}
}
\begin{figure*}[tb]
    \begin{minipage}[t]{0.5\linewidth}
        \includegraphics[width=\linewidth]{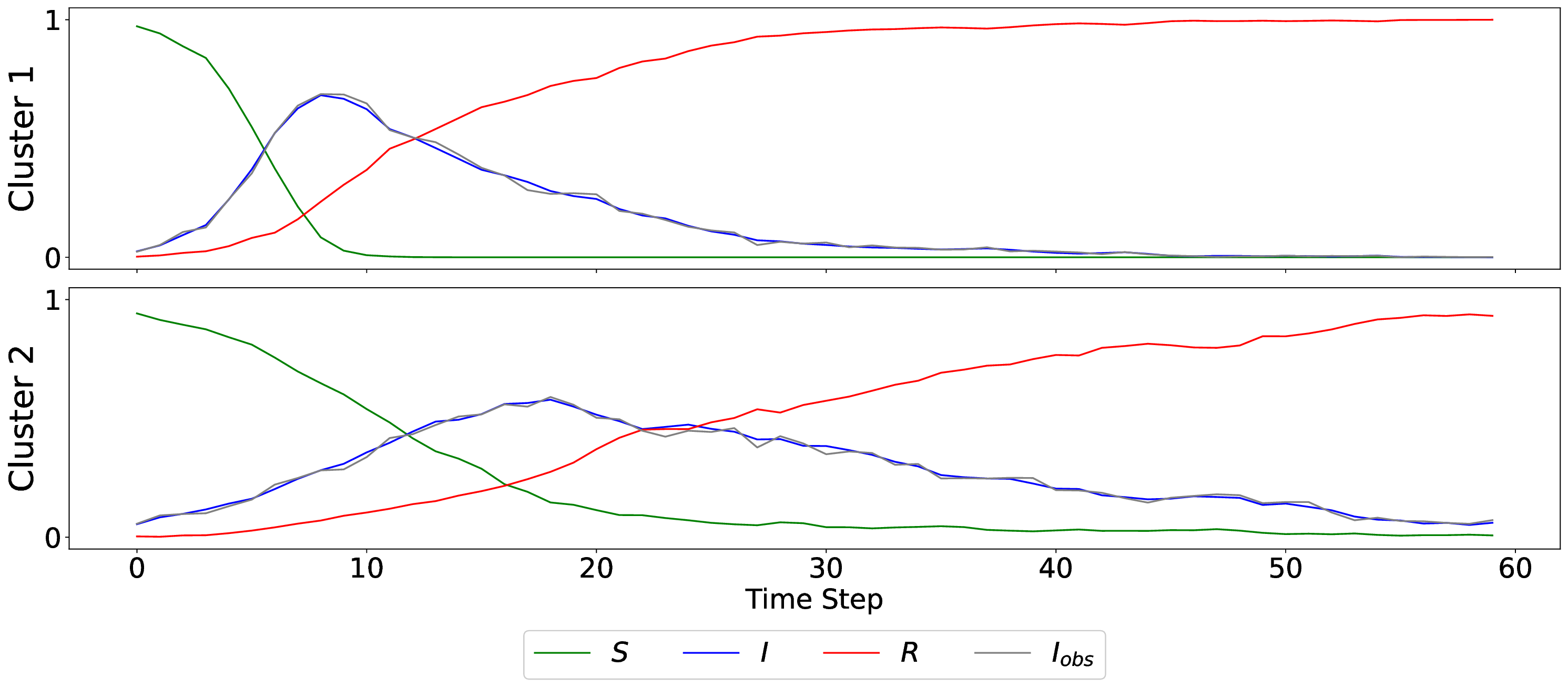}
        \subcaption{Actual time series}
        \label{fig:sir_actual}
    \end{minipage}
    \begin{minipage}[t]{0.5\linewidth}
        \includegraphics[width=\linewidth]{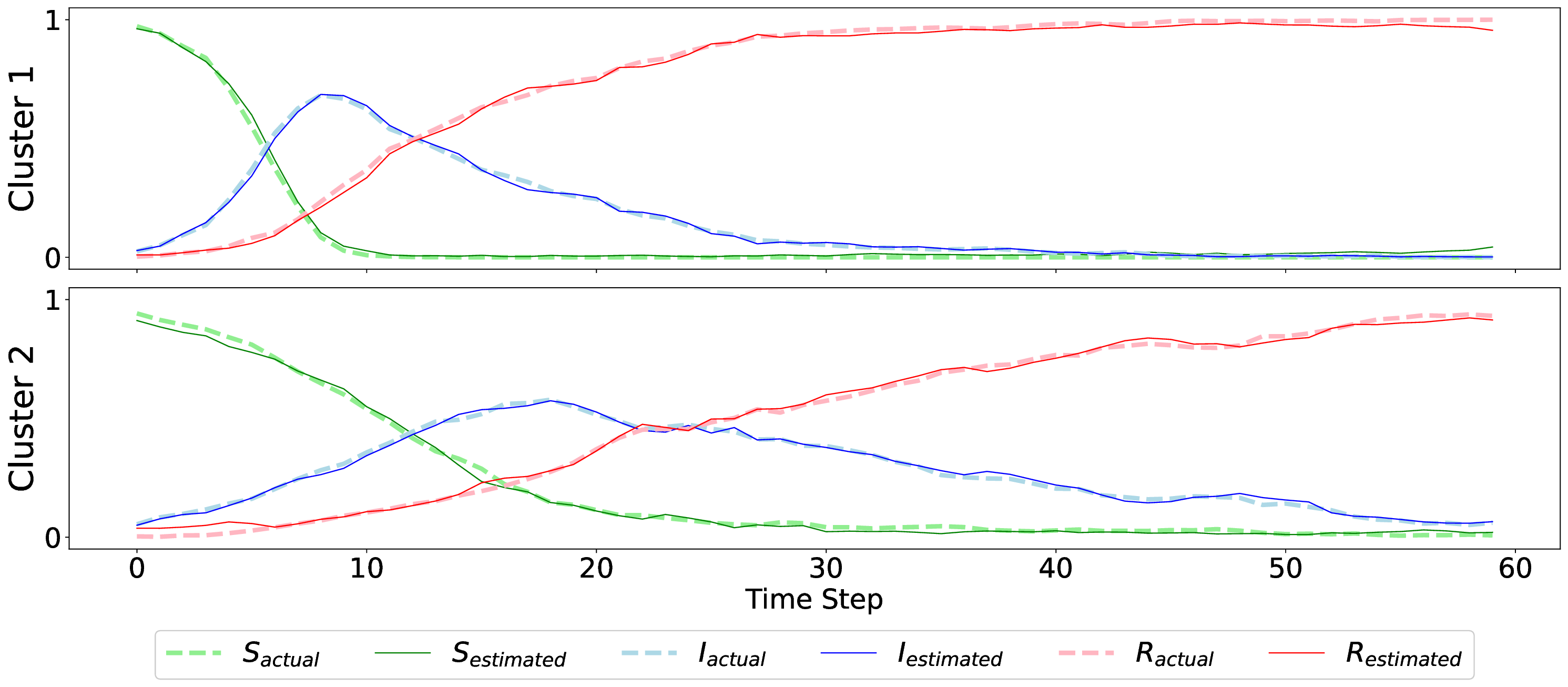}
        \subcaption{Estimated latent variables}
        \label{fig:sir_pred}
    \vspace{0cm}
    \end{minipage}
    %\captionsetup{justification=raggedright,singlelinecheck=false}
    \caption{Representative samples of the actual and estimated time series for the SIR model dataset. (a) Actual time series. The latent variable $I$ is observed as $I_\text{obs}$ under noise. (b) Estimated latent variables. For comparison, the actual latent variables are also plotted.}
    \label{fig:sir}
\end{figure*}
We demonstrate via experiments on simulated datasets that the proposed method is effective for clustering, parameter estimation, and estimating the number of clusters. Specifically, datasets are generated from two dynamics: the Stuart--Landau oscillator \cite{Landau:1944ibh} and the SIR model \cite{kermack1927contribution}. The code is available at \url{https://github.com/ryoichi0917/svi_mssm}.

In our experiments, we use the residual flows \cite{behrmann2019invertible, chen2019residual} as the architecture of the normalizing flows. This is because the residual flows have highly expressive power compared to other normalizing flow architectures, such as MADE \cite{uria2016neurala} and MAF \citep{papamakarios2017masked}.

\subsection{Stuart--Landau Oscillator}

Firstly, we apply the proposed method to time series generated by a discrete-time stochastic version of the Stuart--Landau oscillator. The discrete-time stochastic Stuart--Landau oscillator belonging to the $k$-th cluster is defined as
\begin{subequations}
\begin{align}
x{[t+1]}={}&
    x{[t]}+a_1^{(k)} x{[t]}-b_1^{(k)} y{[t]} \notag\\
    &-\left(x{[t]}^2+y{[t]}^2\right)\left(a_2^{(k)}x{[t]}-b_2^{(k)} y{[t]}\right)
    +\xi_x, \\
y{[t+1]}={}&
    y{[t]}+a_1^{(k)} y{[t]}+b_1^{(k)} x{[t]}\notag\\
    &-\left(x{[t]}^2+y{[t]}^2\right)\left(a_2^{(k)}y{[t]}+b_2^{(k)} x{[t]}\right)
    +\xi_y, \\
y_{\text{obs}}{[t]}={}&y{[t]}+\xi_{\text{obs}},
\end{align}
\end{subequations}
where
\begin{subequations}
\begin{align}
    \begin{bmatrix}
        \xi_{x} & \xi_{y}
    \end{bmatrix}^\top
    &\sim 
    \mathcal{N}\Bigl(0, \mathbf{A}^{(k)}{\mathbf{A}^{(k)}}^\top\Bigr),\\
    \xi_{\text{obs}}
    &\sim 
    \mathcal{N}\Bigl(0, {b^{(k)}}^2\Bigr),\\
    \begin{bmatrix}
        x{[1]} & y{[1]}
    \end{bmatrix}^\top
    &\sim 
    \mathcal{N}\Bigl(
    \begin{bmatrix}
        \mu_{x}^{(k)} &
        \mu_{y}^{(k)} 
    \end{bmatrix}^\top, 
    \mathbf{C^{(k)}}{\mathbf{C}^{(k)}}^\top\Bigr),
\end{align}
\end{subequations}
and $\mathbf{A}^{(k)}$ and $\mathbf{C}^{(k)}$ are lower triangular matrices. 
%In this experiment, $\mathbf{A}^{(k)}, b^{(k)}, \mu_{x}^{(k)}, \mu_{y}^{(k)}, \mathbf{C}^{(k)}$ are common to each cluster, 
To ensure the identifiability of the model, the parameter $a_1^{(k)}$ is assumed to be given. Thus, the parameters to be estimated are $\theta^{(k)}_Q=\{a_2^{(k)}, b_1^{(k)}, b_2^{(k)}, \mathbf{A}^{(k)}\}$, $\theta^{(k)}_R=\{b^{(k)}\}$, and $\theta^{(k)}_{P_0 }=\{\mu_{x}^{(k)}, \mu_{y}^{(k)}, \mathbf{C}^{(k)}\}$, as well as the weights $p^{(k)}$.

The number of clusters was set to three, which means that the dataset is generated from three different Stuart--Landau oscillators. The parameters of each Stuart--Landau oscillator are shown in Table~\ref{tab:sl_params}, and representative samples of the Stuart Landau oscillator dataset are shown in Figure \ref{fig:sl}\subref{fig:sl_actual}. The number of time series belonging to each cluster was 300. Details of the hyperparameters, such as the configuration of the approximate distribution and optimizer settings, are provided in Appendix B.1. 

The values of BIC for the MSSMs with from two to five clusters are shown in Table~\ref{tab:sl}\subref{tab:sl_bic}. 
BIC is the largest when the number of clusters is three, which is the correct number of clusters. Table~\ref{tab:sl}\subref{sl:confusion_matrix} is the confusion matrix that shows the result of clustering. This result indicates that perfect clustering is achieved.

Table~\ref{tab:sl}\subref{tbl:sl_estmated_params} shows the estimated values of the MSSM parameters. From the table, the estimated values are close to the true values for most of the parameters.
Figure~\ref{fig:sl}\subref{fig:sl_pred} shows the actual and estimated latent variables. From the figure, the proposed method accurately estimates the latent variables.

\subsection{SIR Model}

Secondly, we apply the proposed method to time series generated by a discrete-time stochastic 
version of the SIR model \citep{osthus2017forecasting}. The discrete-time stochastic SIR model belonging to the $k$-th cluster is defined as
\begin{subequations}
\label{eq:SIR}
\begin{align}
    \bm{\rho}[t+1] \mid \bm{\rho}[t]
        &\sim \text{Dirichlet}(10^{{a}^{(k)}}
             \bm{f}^{(k)}(\bm{\rho}[t])),\\
    I_{\text{obs}}[t] \mid \rho_{(I)}[t] 
        &\sim \text{Beta}(10^{{b}^{(k)}} \rho_{(I)}[t] , 
                   10^{{b}^{(k)}} (1-\rho_{(I)}[t] )),\\
    \bm{\rho}[1]
        &\sim \text{Dirichlet}(10^{{c}^{(k)}}\bm{\rho}_{(\text{init})}^{(k)}) ,
\end{align}
\end{subequations}
where 
\begin{subequations}
\begin{gather}
\bm{f}^{(k)}(\bm{\rho}[t]) = 
    \begin{bmatrix}
    f_{(S)}^{(k)}(\bm{\rho}[t]), 
    f_{(I)}^{(k)}(\bm{\rho}[t]), 
    f_{(R)}^{(k)}(\bm{\rho}[t])
    \end{bmatrix}^\top,\\
\left.
\begin{aligned}   
    f_{(S)}^{(k)}(\bm{\rho}[t])&= \rho_{(S)}[t] 
        - \beta^{(k)} \rho_{(S)}[t]\rho_{(I)}[t],\\
    f_{(I)}^{(k)}(\bm{\rho}[t])&=\rho_{(I)}[t] 
        +\beta^{(k)}\rho_{(S)}[t]\rho_{(I)}[t] -\gamma^{(k)}\rho_{(I)}[t] ,\\
    f_{(R)}^{(k)}(\bm{\rho}[t])&=\rho_{(R)}[t]+\gamma^{(k)}\rho_{(I)}[t],
\end{aligned}\quad \right\}\\
\bm{\rho}{[t]} = 
    \begin{bmatrix}
    \rho_{(S)}[t], \rho_{(I)}[t], \rho_{(R)}[t]
    \end{bmatrix}^\top,\\
\bm{\rho}_{(\text{init})}^{(k)} = 
    \begin{bmatrix}
    \rho_{(S, \text{init})}^{(k)}, \rho_{(I, \text{init})}^{(k)}, \rho_{(R, \text{init})}^{(k)}
    \end{bmatrix}^\top,\\
\rho_{(S, \text{init})}^{(k)} + \rho_{(I, \text{init})}^{(k)}  +\rho_{(R, \text{init})}^{(k)}=1.
\end{gather}
\end{subequations}
Thus, the parameters to be estimated are  
 $\theta^{(k)}_Q=\{\beta^{(k)}, \gamma^{(k)}, a^{(k)}\}$, $\theta^{(k)}_R=\{b^{(k)}\}$, and
$\theta^{(k)}_{P_0}=\{c^{(k)}, \bm{\rho}_{(init)}^{(k)}\}$, as well as the weights $p^{(k)}$.
Although the SIR model (\ref{eq:SIR}) is intractable for normalizing flows due to the Dirichlet distributions, we can apply the proposed method by rewriting Eq.~(\ref{eq:SIR}) into an equivalent form. See Appendix B.2 for details.

The number of clusters was set to two, which means that the dataset is generated from two different SIR models. The parameters of each SIR model are shown in Table~\ref{tab:sir_params}, and representative samples of the SIR model dataset are shown in Figure~\ref{fig:sir}\subref{fig:sir_actual}. The number of time series belonging to each cluster was 300. Details of the hyperparameters, such as the configuration of the approximate distribution and optimizer settings, are provided in Appendix B.1. 

The values of BIC for the MSSMs with from two to four clusters are shown in Table~\ref{tab:sir}\subref{tab:sir_bic}. BIC is the largest when the number of clusters is two, which is the correct number of clusters. Table~\ref{tab:sir}\subref{tab:sir_confusion_matrix} is the confusion matrix that shows the result of clustering. This result indicates that perfect clustering is achieved.

Table~\ref{tab:sir}\subref{tab:sir_estmated_params} shows the estimated values of the MSSM parameters. From the table, the estimated values are close to the true values for most of the parameters.
Figure~\ref{fig:sir}\subref{fig:sir_pred} shows the actual and estimated latent variables. From the figure, the proposed method accurately estimates the latent variables.

\section{Conclusion}
In this paper, we propose a novel method of model-based time series clustering with mixtures of general state-space models. An advantage of the proposed method is that it enables the use of tailored time series models. This not only improves clustering and prediction accuracy but also enhances the interpretability of the estimated parameters. Experiments on simulated datasets show that the proposed method is effective for clustering, parameter estimation, and estimating the number of clusters.

A limitation of this study is that the validation experiments in this paper were only conducted on simulated datasets. Future research should validate the effectiveness of the proposed method on real datasets. When applying the proposed method to real data, it would also be useful to enhance the structure of the MSSMs to allow for the input of exogenous variables.

\section*{Acknowledgments}
This work was supported by Osaka Gas Co., Ltd.

%\bibliographystyle{elsarticle-num}
%\bibliography{refs_2}

%%%%%%%%%% Prefix a "S" to all references %%%%%%%%%%
\renewcommand{\bibnumfmt}[1]{[a#1]}
\renewcommand{\citenumfont}[1]{a#1}
%%%%%%%%%% Prefix a "S" to all references %%%%%%%%%%

\clearpage

\appendix
\section{Practical Efforts for Parameter Optimization}
\label{append:param optim}

In this section, we describe three tricks to avoid undesirable convergence of $q_{\phi}(z_i \mid \mathbf{Y}_i)$ to a local optimum, other than entropy annealing.

The first of these tricks relates to the optimizer. In the proposed method, (simple) mini-batch SGD is used to optimize MSSMs. This is because mini-batch SGD can accurately approximate the posterior distribution in SVI \cite{m2017stochastic}. The learning rate of mini-batch SGD is changed periodically based on cyclical learning rates \cite{smith2017cyclical} to avoid convergence to a local optimum.

The second is to introduce learning rate scheduling and fine-tuning. When MSSMs are trained with cyclical learning rates, the parameters do not converge, because the learning rate increases periodically. Therefore, we set the learning rate to a low value at the late stages of the training to make the parameters converge. After this training, fine-tuning is performed to learn only $\theta^{(k)}_{P_0}$ and parameters of normalizing flow modules. This is because $\theta^{(k)}_{P_0}$ is poorly optimized if all parameters are simultaneously learned. The loss function is significantly less affected by the improvement of $\theta^{(k)}_{P_0}$ than by that of $\theta^{(k)}_{R}$ due to the fact
\begin{align}
&\log p_{\mathbf{\theta}_{P_0^{(k)}},
        \mathbf{\theta}_{R^{(k)}}}(\mathbf{X}_i \mid z_i=k)\notag\\
&= \log p_{
    \mathbf{\theta}_{P_0^{(k)}}}
        (\mathbf{x}_i[t] \mid z_i=k)
    + \sum_{t=2}^T
        \log p_{
        \mathbf{\theta}_{R^{(k)}}}
        (\mathbf{x}_i[t+1] \mid \mathbf{x}_i[t], z_i=k).
\end{align}

Finally, we train MSSMs multiple times from different initial parameter values and adopt the model with the lowest loss to avoid rare cases unsolved by the above-mentioned two tricks.

With the above tools and entropy annealing, MSSMs converged without falling into local optimum solutions in the experiments.

\section{Details of Experiments}

\subsection{Model Architecture}
\begin{figure}[t]
    \centering
    \footnotesize
    \includegraphics[width=0.5\linewidth]{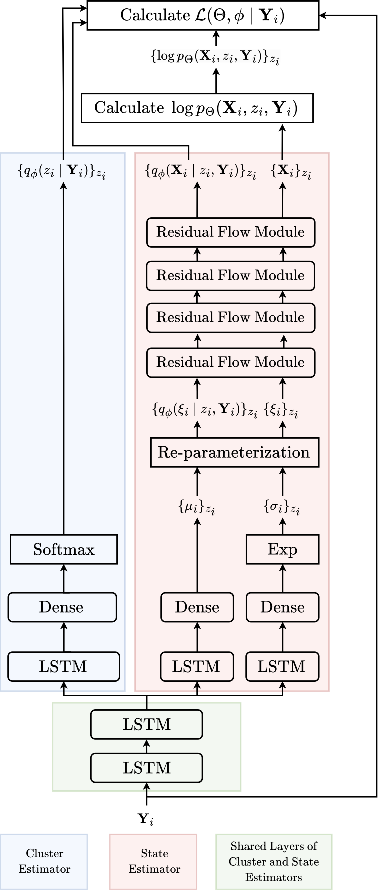}
    \caption{Details of model architecture.}
    \label{fig:model_detail}
\end{figure}
The architectural details of the model used in our experiments are shown in Figure~\ref{fig:model_detail}. For the hyperparameter settings of each layer, see the code on GitHub (\url{https://github.com/ryoichi0917/svi_mssm}).

As shown in Figure~\ref{fig:model_detail}, we used the conventional LSTM-based architecture to clearly demonstrate the fundamental concept of the proposed method. By using more efficient architectures based on, for example, quasi-recurrent neural networks \cite{bradbury2022quasirecurrent} or Legendre memory units \cite{chilkuri2021parallelizing}, the computational cost can be reduced further.

\label{append:Experiments}

\subsection{Equivalent form of the SIR Model}
\label{append:eq_form_of_SIR}
For the stochastic SIR model, the representation (22) is intractable for normalizing flows due to the Dirichlet distribution. Nevertheless, we can apply the proposed method by rewriting the model in the equivalent form as
 \begin{subequations}
\begin{gather}
\left.
\begin{aligned}   
    G_{(S)}[t+1]\sim \text{Gamma} (10^{{a}^{(k)}} f_{(S)}^{(k)}(\bm{\rho}[t]), 1),\\
    G_{(I)}[t+1]\sim \text{Gamma} (10^{{a}^{(k)}} f_{(I)}^{(k)}(\bm{\rho}[t]), 1),\\
    G_{(R)}[t+1]\sim \text{Gamma} (10^{{a}^{(k)}} f_{(R)}^{(k)}(\bm{\rho}[t]), 1),
\end{aligned}\quad \right\}\\
    I_{\text{obs}}[t] \mid \rho_{(I)}[t]
        \sim \text{Beta}(
            10^{{b}^{(k)}}\rho_{(I)}[t] , 
            10^{{b}^{(k)}}(1-\rho_{(I)}[t])),\\
\left.
\begin{aligned}  
    G_{(S)}[1]&\sim \text{Gamma} (10^{{c}^{(k)}} \rho_{(S, \text{init})}^{(k)}, 1),\\
    G_{(I)}[1]&\sim \text{Gamma} (10^{{c}^{(k)}} \rho_{(I, \text{init})}^{(k)}, 1),\\
    G_{(R)}[1]&\sim \text{Gamma} (10^{{c}^{(k)}} \rho_{(R, \text{init})}^{(k)}, 1),
\end{aligned}\quad \right\}
\end{gather}
\end{subequations}
where
\begin{subequations}
\begin{gather}
\left.
\begin{aligned}   
    f_{(S)}^{(k)}(\bm{\rho}[t])&= \rho_{(S)}[t] 
        - \beta^{(k)} \rho_{(S)}[t]\rho_{(I)}[t],\\
    f_{(I)}^{(k)}(\bm{\rho}[t])&=\rho_{(I)}[t] 
        +\beta^{(k)}\rho_{(S)}[t]\rho_{(I)}[t] -\gamma^{(k)}\rho_{(I)}[t] ,\\
    f_{(R)}^{(k)}(\bm{\rho}[t])&=\rho_{(R)}[t]+\gamma^{(k)}\rho_{(I)}[t],
\end{aligned}\quad \right\}\\
\left.
\begin{aligned}
    \rho_{(S)}[t]&=\frac{G_{(S)}[t]}
                                {G_{(S)}[t]+G_{(I)}[t]+G_{(R)}[t]},\\
    \rho_{(I)}[t]&=\frac{G_{(I)}[t]}
                                {G_{(S)}[t]+G_{(I)}[t]+G_{(R)}[t]},\\
    \rho_{(R)}[t]&=\frac{G_{(R)}[t]}
                                {G_{(S)}[t]+G_{(I)}[t]+G_{(R)}[t]},
\end{aligned}\quad \right\}\\
    \bm{\rho}{[t]} = 
        \begin{bmatrix}
        \rho_{(S)}[t], \rho_{(I)}[t],\rho_{(R)}[t]
        \end{bmatrix}^\top,\\
    \rho_{(S, \text{init})}^{(k)} + \rho_{(I, \text{init})}^{(k)}  +\rho_{(R, \text{init})}^{(k)}=1.
\end{gather}
\end{subequations}

\end{document}